%% file: egpaper_for_review.tex
\documentclass[10pt,twocolumn,letterpaper]{article}

\usepackage{cvpr}
\usepackage{times}
\usepackage{epsfig}
\usepackage{graphicx}
\usepackage{amsmath}
\usepackage{amssymb}

\usepackage[pagebackref=true,breaklinks=true,letterpaper=true,colorlinks,bookmarks=false]{hyperref}

\cvprfinalcopy % *** Uncomment this line for the final submission

\def\cvprPaperID{****} % *** Enter the CVPR Paper ID here

\ifcvprfinal\fi
\begin{document}

%%%%%%%%% TITLE
\title{On the Robustness of Monte Carlo Dropout Trained with Noisy Labels}

\author{Purvi Goel\\
Facebook\\
{\tt\small pgoel2@fb.com}
\and
Li Chen\\
Facebook\\
{\tt\small lichen66@fb.com}
}

\maketitle
%\thispagestyle{empty}

%%%%%%%%% ABSTRACT
\begin{abstract}

   The memorization effect of deep learning hinders its performance to effectively generalize on test set when learning with noisy labels. Prior study has discovered that epistemic uncertainty techniques are robust when trained with noisy labels compared with neural networks without uncertainty estimation. They obtain prolonged memorization effect and better generalization performance under the adversarial setting of noisy labels. Due to its superior performance amongst other selected epistemic uncertainty methods under noisy labels, we focus on Monte Carlo Dropout (MCDropout) and investigate why it is robust when trained with noisy labels. Through empirical studies on datasets MNIST, CIFAR-10, Animal-10n, we deep dive into three aspects of MCDropout under noisy label setting: 1. efficacy: understanding the learning behavior and test accuracy of MCDropout when training set contains artificially generated or naturally embedded label noise; 2. representation volatility: studying the responsiveness of neurons by examining the mean and standard deviation on each neuron's activation; 3. network sparsity: investigating the network support of MCDropout in comparison with deterministic neural networks. %While in general, adding dropout layers sparsifies deep neural networks, 
   Our findings suggest that MCDropout further sparsifies and regularizes the deterministic neural networks and thus provides higher robustness against noisy labels. 
 
\end{abstract}

%%%%%%%%% BODY TEXT
\section{Introduction}

Neural networks exhibit state-of-the-art performance on many learning tasks, such as classification and segmentation. However, training these networks requires an abundance of carefully labeled data; networks tend to overfit quickly to noise in training labels, which makes their application to noisy real-world problems less effective. Expert-labeled data is expensive and time-consuming to collect; label noise is common in less carefully crafted datasets due to measurement inaccuracies, human error, etc. 

Nonetheless the latter type of data, albeit noisy, is much more readily available and in much larger quantities. One recent strategy shown to perform well on datasets containing significant amounts of label noise is augmenting the neural network with an uncertainty estimation method like Monte Carlo Dropout \cite{chen2021}. These uncertainty estimation models display a delayed memorization effect of noisy training labels, and can generalize better to clean test data. Augmenting models with Monte Carlo Dropout shows a slower degradation of classification performance, consistent on benchmark datasets like MNIST and CIFAR-10\cite{chen2021}. 
In addition to its resilient performance against noisy training labels, MCDropout also does not add training overhead and only adds minimal cost to inference time. 

The robustness property and low-computational cost of MCDropout indicate it as an effective and practical solution against noisy labels. In this paper, our goal is to not only determine whether MCDropout performs consistently better in these noisy-label situations, but also provide an in-depth analysis for \emph{why} it performs better. We present an investigation into the performance and latent representation learned by a model augmented with MCDropout. We first evaluate the accuracy of MCDropout models in comparison with deterministic neural networks on datasets like MNIST and CIFAR-10 with artificially injected noisy labels and Animal-10n with natural annotation noise. Second, we measure neuron responsiveness in each layer, to better explore the differences between latent representations learned by certainty and MCDropout models. Finally we study network sparsity and find that the sparsity property offered by MCDropout models contribute to robustness against noisy training labels. To our knowledge, our work provides the first detailed analysis of MCDropout in the setting of noisy labels. 

The rest of the paper is organized as follows. In Section \ref{sec:prelim}, we provide the background information on the noisy label setting, label noise taxonomies, Monte Carlo Dropout and related work.  In Section \ref{sec:method}, we describe our study directions including measuring efficacy, neuron responsiveness via volatility and network sparsity.  In Section \ref{sec:experiments}, we demonstrate the effectiveness of MCDropout on empirical datasets such as MNIST, CIFAR10 with artificially corrupted training labels and Animal-10n a real-world dataset containing annotation noise. We further analyze the neuron responsiveness and network sparsity by MCDropout in comparison with deterministic networks. Finally in Section \ref{sec:discussion}, we discuss optimal placement for MCDropout on a neural network and conclude our paper.

 % \textcolor{red}{thoughts (Li): 1. stress we are the first one to do this. 2. might want to add a motivational image from prior work after Section 2.    thoughts: add a motivation section to quote one img from wiml paper?
% }
 
% Finally, we show applications of these findings in tuning and hyperparameter selection for MCDropout models. We use our results to form intuition on the best neural network layers to convert to MCDropout models.

%------------------------------------------------------------------------
\input{latex/prelim}

\begin{figure*}[t!]
    \centering
    \setlength{\tabcolsep}{1pt}
    \begin{tabular}{cc c cc c cc c cc c cc}
      \rotatebox{90}{Certain}
      \includegraphics[width=0.09\linewidth]{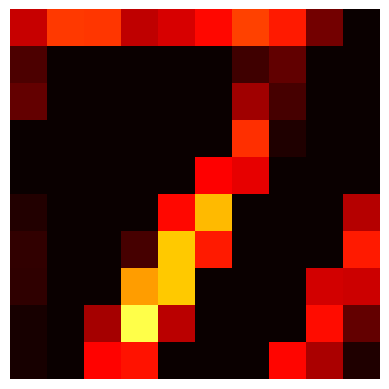} &
      \includegraphics[width=0.09\linewidth]{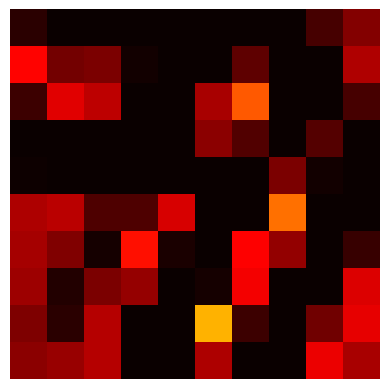} &
      &
      \includegraphics[width=0.09\linewidth]{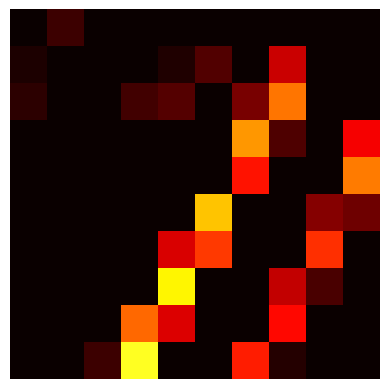} &
      \includegraphics[width=0.09\linewidth]{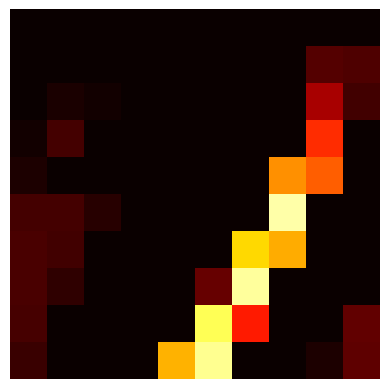} &
      &
      \includegraphics[width=0.09\linewidth]{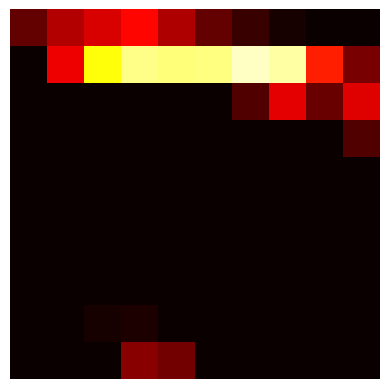} &
      \includegraphics[width=0.09\linewidth]{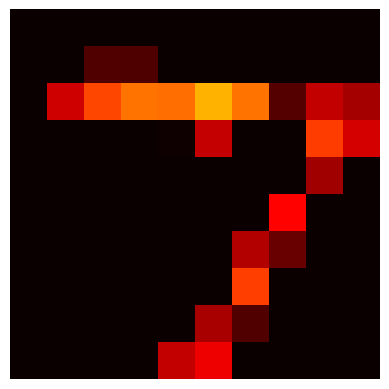} &
      &
      \includegraphics[width=0.09\linewidth]{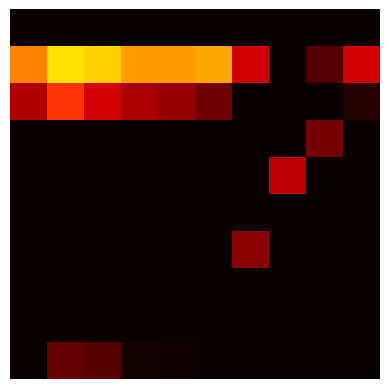} &
      \includegraphics[width=0.09\linewidth]{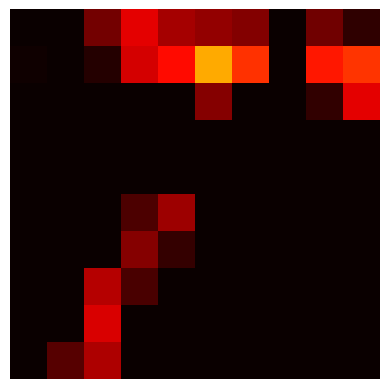} &
      &
      \includegraphics[width=0.09\linewidth]{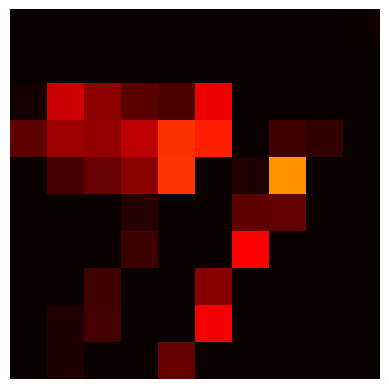} &
      \includegraphics[width=0.09\linewidth]{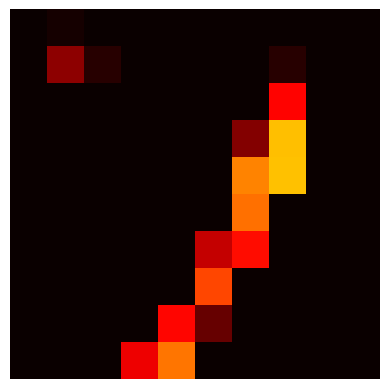}
      \\
      \rotatebox{90}{MCDropout}
      \includegraphics[width=0.09\linewidth]{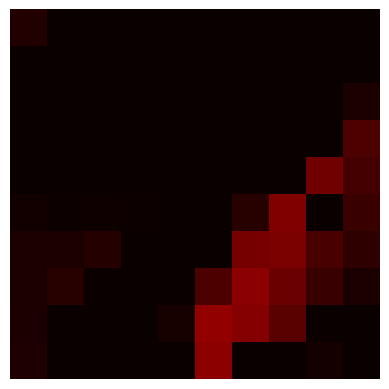} &
      \includegraphics[width=0.09\linewidth]{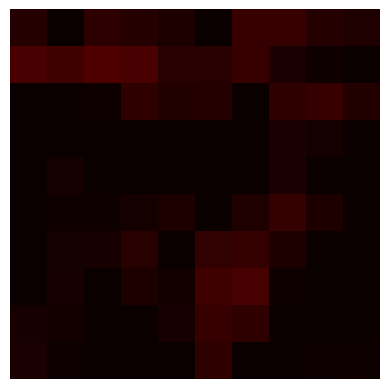} &
      &
      \includegraphics[width=0.09\linewidth]{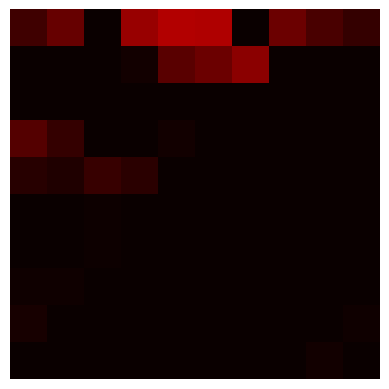} &
      \includegraphics[width=0.09\linewidth]{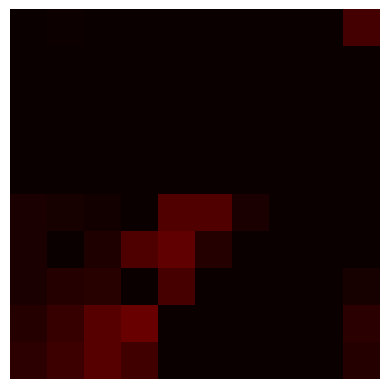} &
      &
      \includegraphics[width=0.09\linewidth]{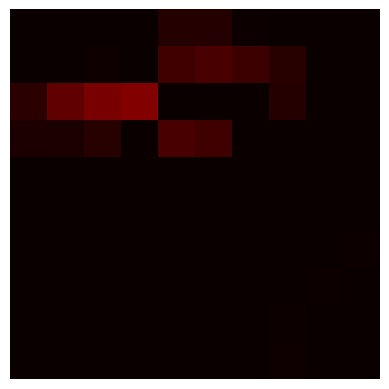} &
      \includegraphics[width=0.09\linewidth]{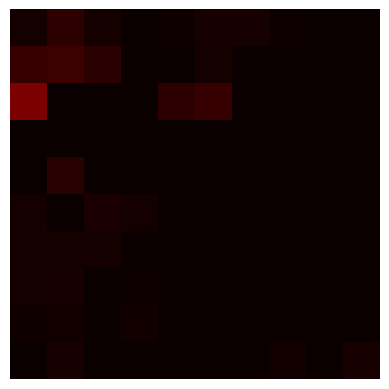} &
      &
      \includegraphics[width=0.09\linewidth]{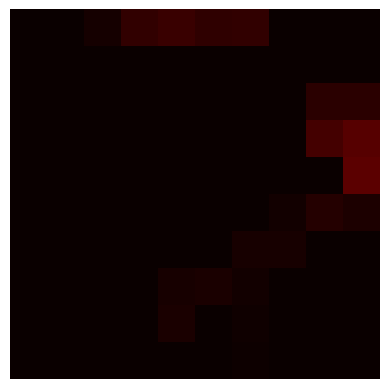} &
      \includegraphics[width=0.09\linewidth]{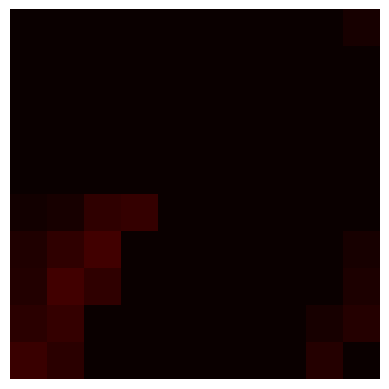} &
      &
      \includegraphics[width=0.09\linewidth]{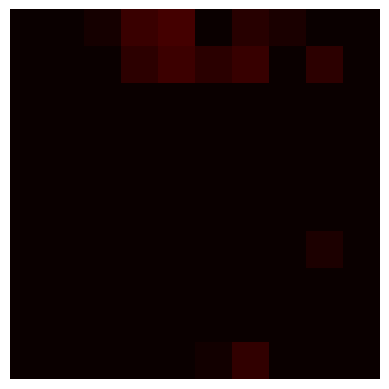} &
      \includegraphics[width=0.09\linewidth]{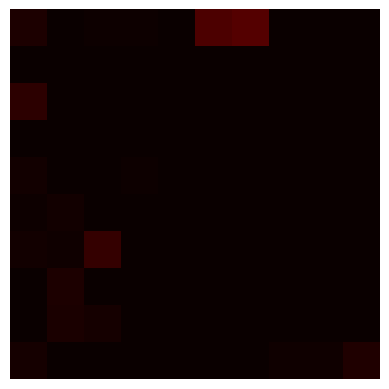}
      \\
  \end{tabular}
    \caption{
    Activation maps for our MNIST experiment. We show the activations from 10 neurons on a random image from MNIST's test set, from the certain LeNet5's second convolution layer (top) and the MCDropout model's second convolution layer. Brighter values indicate higher activation values. The chosen layer contains 16 neurons total; for each model, we choose the 10 neurons with the highest mean activation value. The certain model activates much more strongly, while the MCDropout model's activations are more muted, and spatially sparse. }
    \label{fig:heatmap_mnist}
\end{figure*}

\begin{figure*}[t!]
    \centering
    \setlength{\tabcolsep}{1pt}
    \begin{tabular}{cc c cc c cc c cc c cc}
      \rotatebox{90}{Certain}
      \includegraphics[width=0.09\linewidth]{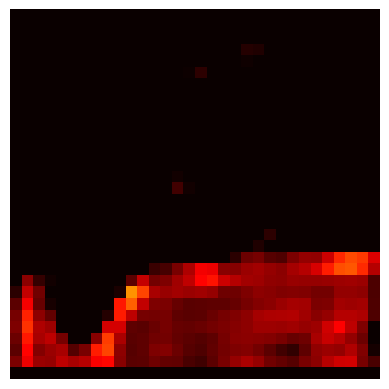} &
      \includegraphics[width=0.09\linewidth]{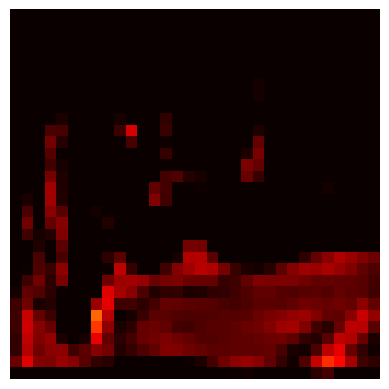} &
      &
      \includegraphics[width=0.09\linewidth]{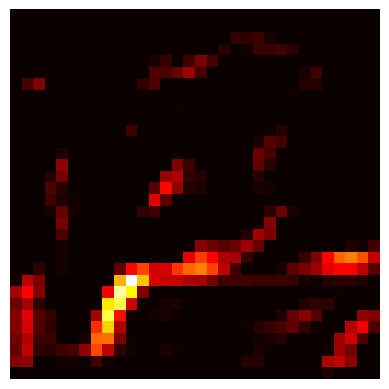} &
      \includegraphics[width=0.09\linewidth]{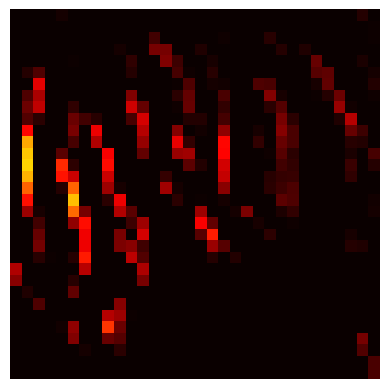} &
      &
      \includegraphics[width=0.09\linewidth]{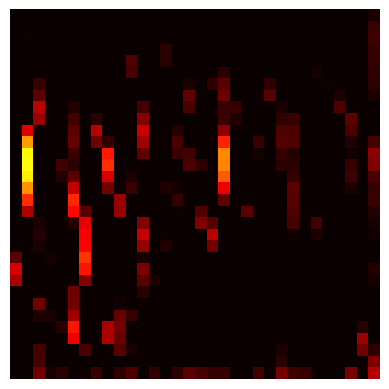} &
      \includegraphics[width=0.09\linewidth]{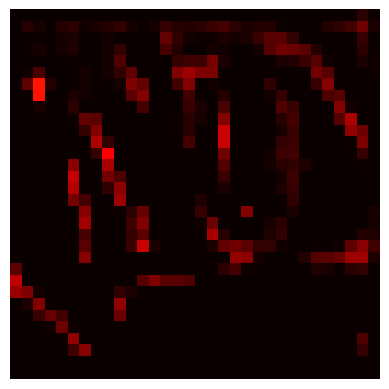} &
      &
      \includegraphics[width=0.09\linewidth]{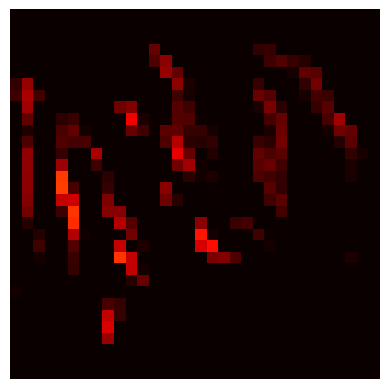} &
      \includegraphics[width=0.09\linewidth]{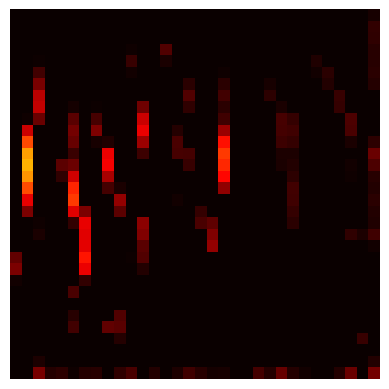} &
      &
      \includegraphics[width=0.09\linewidth]{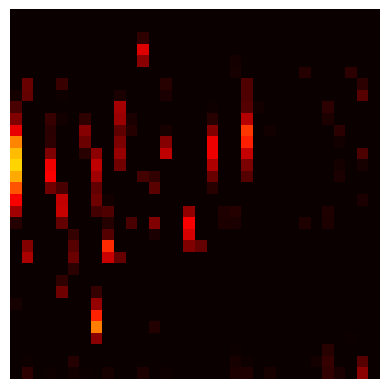} &
      \includegraphics[width=0.09\linewidth]{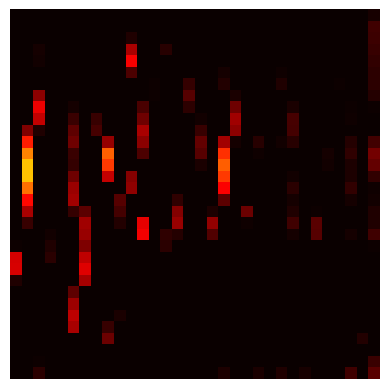}
      \\
      \rotatebox{90}{MCDropout}
      \includegraphics[width=0.09\linewidth]{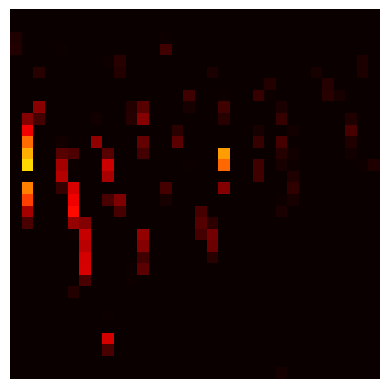} &
      \includegraphics[width=0.09\linewidth]{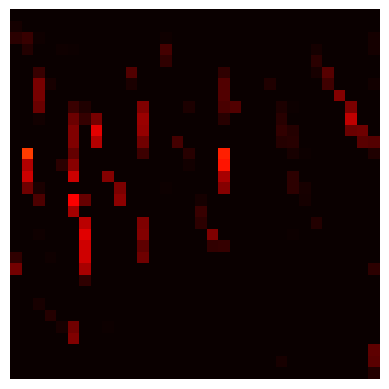} &
      &
      \includegraphics[width=0.09\linewidth]{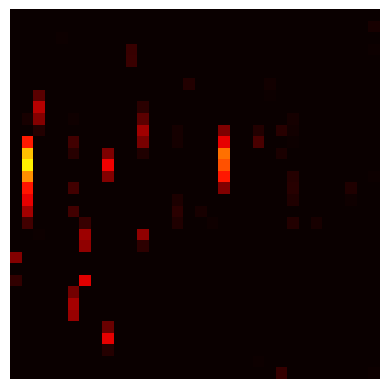} &
      \includegraphics[width=0.09\linewidth]{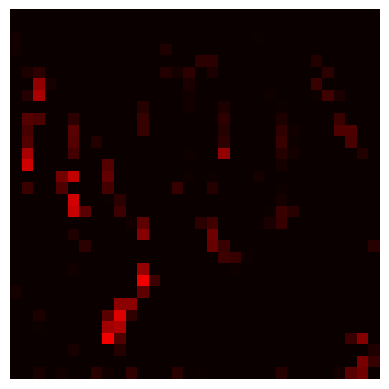} &
      &
      \includegraphics[width=0.09\linewidth]{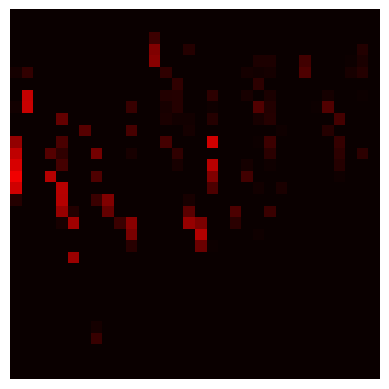} &
      \includegraphics[width=0.09\linewidth]{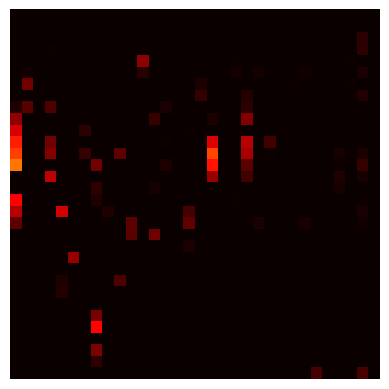} &
      &
      \includegraphics[width=0.09\linewidth]{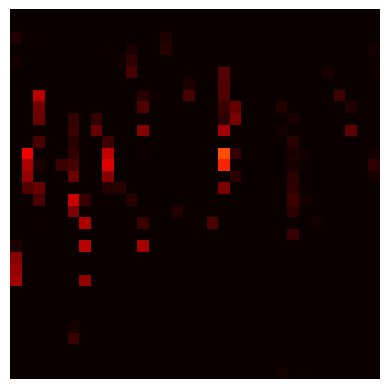} &
      \includegraphics[width=0.09\linewidth]{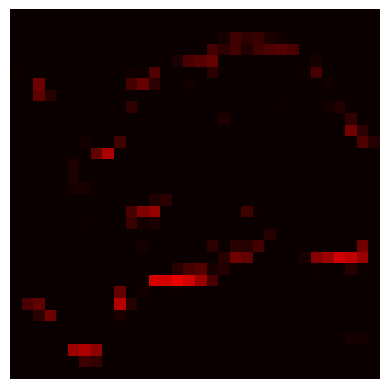} &
      &
      \includegraphics[width=0.09\linewidth]{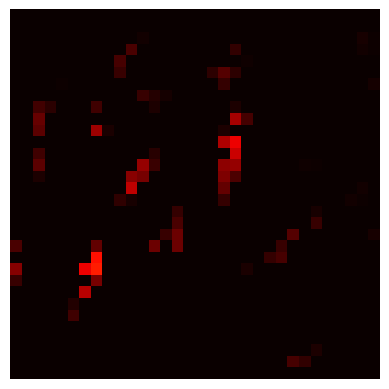} &
      \includegraphics[width=0.09\linewidth]{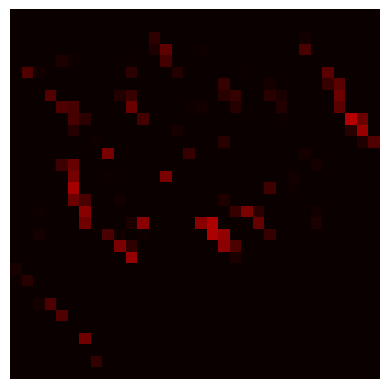}
      \\
  \end{tabular}
    \caption{Activation maps for our CIFAR10 experiment. We show the activations from 10 neurons on a random image from CIFAR10's test set, from the certain ConvNet's first convolution layer (top) and the MCDropout model's first convolution layer. The chosen layer contains 48 neurons total; for each model, we choose the 10 neurons with the highest mean activation value. The brighter the value at a pixel, the higher the activation at that point. We see on visual inspection that MCDropout has smaller overall activations: there are no bright yellow areas in the bottom row's feature maps, while we do for several maps in the top row. In addition, the activations are more spatially sparse: many of the displayed MCDropout model's featuremaps have entire regions with no activations. This is not as noticeable in the certain model. }
    \label{fig:heatmap_cifar}
\end{figure*}

\begin{figure*}[t!]
    \centering
    \setlength{\tabcolsep}{1pt}
    \begin{tabular}{cc c cc c cc c cc c cc}
      \rotatebox{90}{Certain}
      \includegraphics[width=0.09\linewidth]{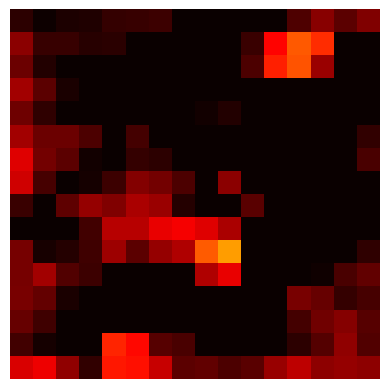} &
      \includegraphics[width=0.09\linewidth]{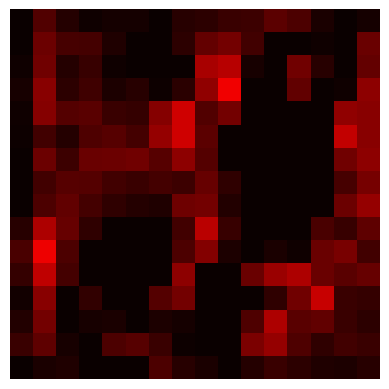} &
      &
      \includegraphics[width=0.09\linewidth]{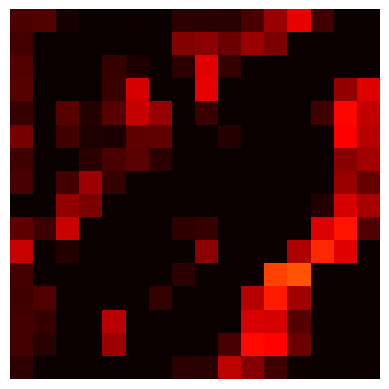} &
      \includegraphics[width=0.09\linewidth]{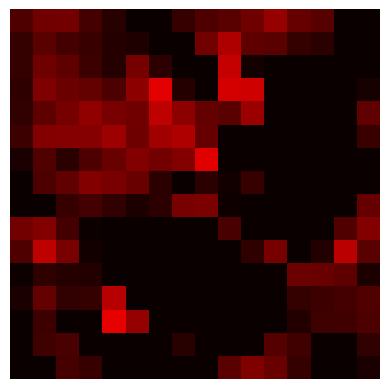} &
      &
      \includegraphics[width=0.09\linewidth]{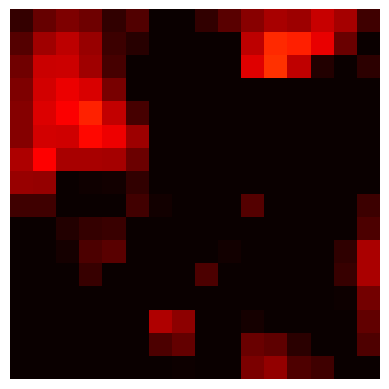} &
      \includegraphics[width=0.09\linewidth]{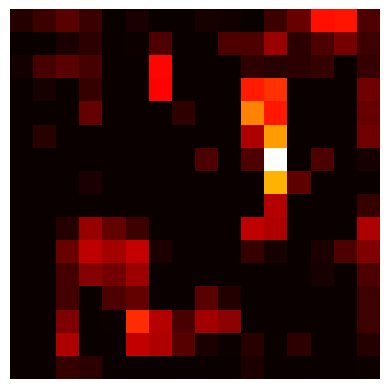} &
      &
      \includegraphics[width=0.09\linewidth]{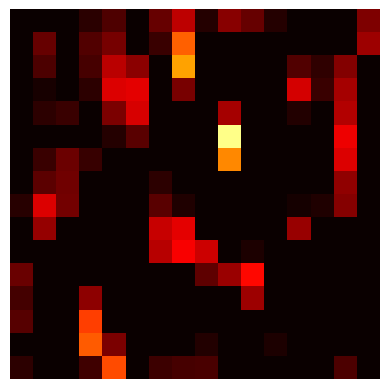} &
      \includegraphics[width=0.09\linewidth]{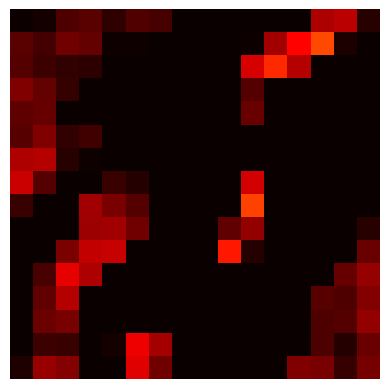} &
      &
      \includegraphics[width=0.09\linewidth]{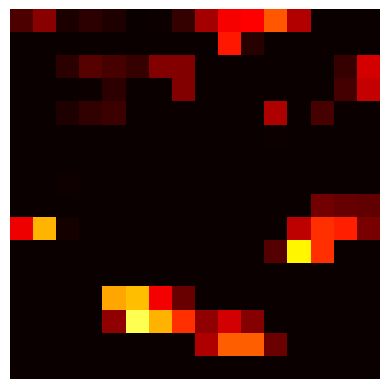} &
      \includegraphics[width=0.09\linewidth]{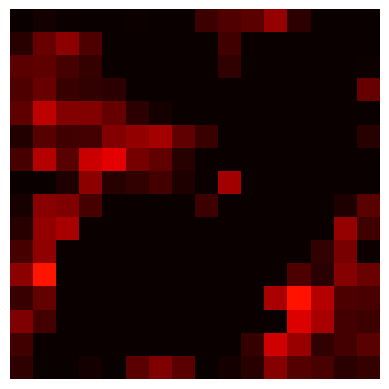}
      \\
      \rotatebox{90}{MCDropout}
      \includegraphics[width=0.09\linewidth]{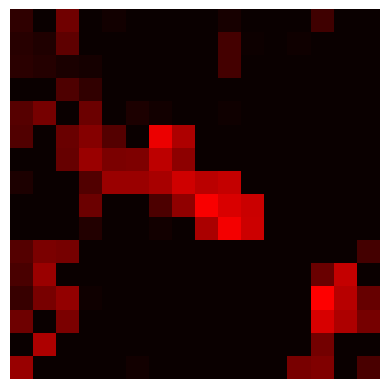} &
      \includegraphics[width=0.09\linewidth]{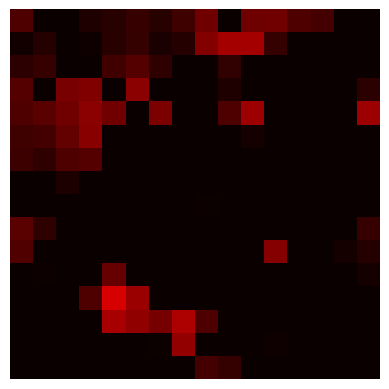} &
      &
      \includegraphics[width=0.09\linewidth]{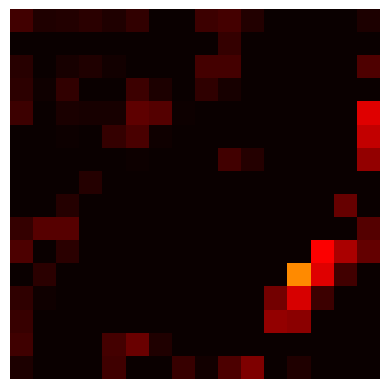} &
      \includegraphics[width=0.09\linewidth]{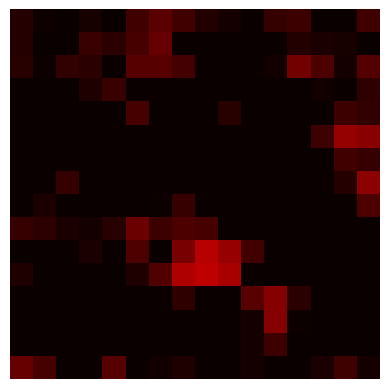} &
      &
      \includegraphics[width=0.09\linewidth]{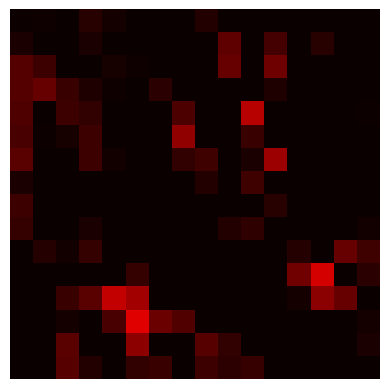} &
      \includegraphics[width=0.09\linewidth]{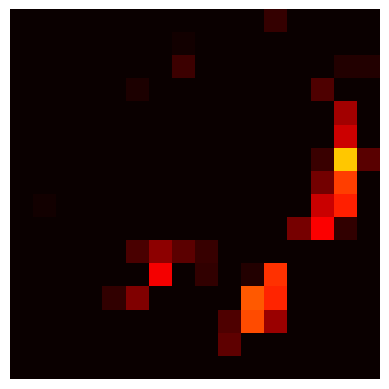} &
      &
      \includegraphics[width=0.09\linewidth]{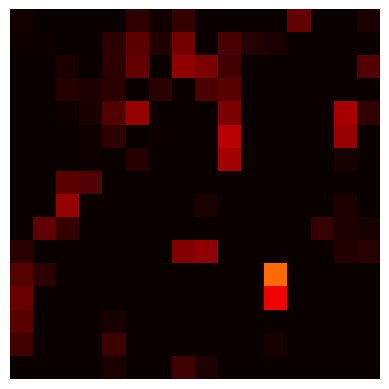} &
      \includegraphics[width=0.09\linewidth]{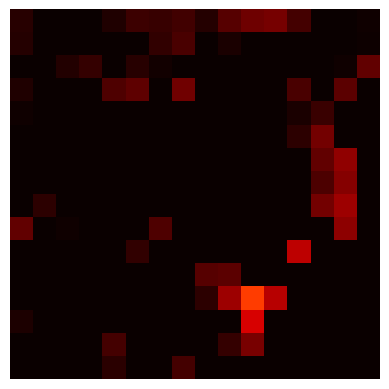} &
      &
      \includegraphics[width=0.09\linewidth]{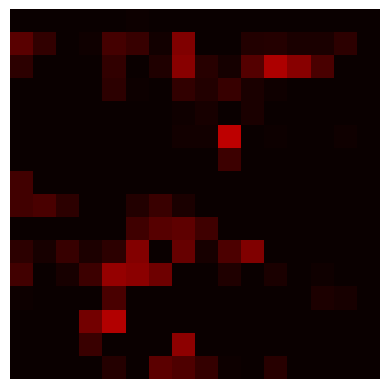} &
      \includegraphics[width=0.09\linewidth]{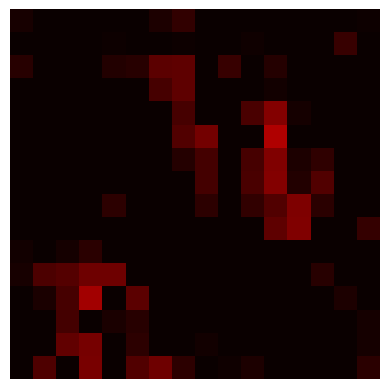}
      \\
  \end{tabular}
    \caption{
    Activation maps for our Animal-10n experiment. We show the activations from 10 neurons on a random image from Animal-10n's test set, from the certain ConvNet's third convolution layer (top) and the MCDropout model's third convolution layer. Brighter values indicate higher activation values. The chosen layer contains 196 neurons total; for each model, we choose the 10 neurons with the highest mean activation value. While less obvious than the comparison in Figure \ref{fig:heatmap_cifar} and Figure \ref{fig:heatmap_mnist}, the certain model still shows higher and less spatially sparse activations than the MCDropout model.}
    \label{fig:heatmap_animal10}
\end{figure*}

\section{Investigation} \label{sec:method}

%\textcolor{red}{thoughts: should we merge section 3 and 4? if we leave section 3 as independent, let's revise the subsections within 3, 4 so they are not redundant.}
Our goal is to analyze the latent representations learned by MCDropout models, particularly in comparison with certainty models, trained in the presence of noisy labels. Similar to the definitions presented in Bau et al~\cite{bau2019gandissect}, we use the term \emph{representation} to describe the outputs of a particular layer in a model. More specifically: which channels of the layer have been activated for various data inputs? How strongly have these channels been activated? What is the variation in a specific channel's possible activations? Comparing the representations lends insight and intuition as to why one model may perform better than another one. Essentially, we investigate why MCDropout performs better than a certainty model by comparing the different latent representations learned by the two models respectively.

Again following the vocabulary used in Bau et al~\cite{bau2019gandissect}, we refer to feature maps as the output of every layer in the network--the  aggregate of the feature maps makes up the network's learned representation. 
We refer to a neuron as a specific channel of the feature map. In this paper, we use the term \emph{activation gamut} to refer to all the possible values that a particular neuron can produce. We can approximate the activation gamut as the set of a neuron's activation values for each image in a dataset. 

We compare the classification efficacy, neuron responsiveness, and network sparsity by the two models respectively. 
To understand how two models have learned and encoded information differently, we evaluate trained models on test set and cache neuron activations from each layer, where we derive statistics such as mean and standard deviations on each neuron with respect to data samples from the test set. % We compare statistics of the activations between the models along three axes: classification efficacy, neuron responsiveness, and network sparsity. 

\begin{figure*}[t!]
    \centering
    \setlength{\tabcolsep}{1pt}
    \begin{tabular}{ccc }
      
      \includegraphics[width=0.3\linewidth]{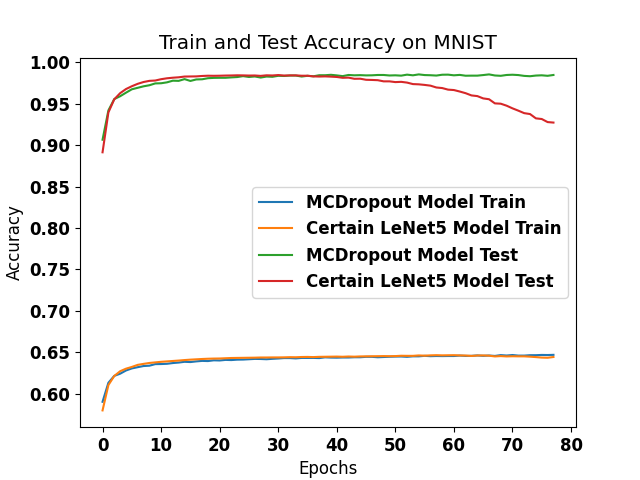} &
      \includegraphics[width=0.3\linewidth]{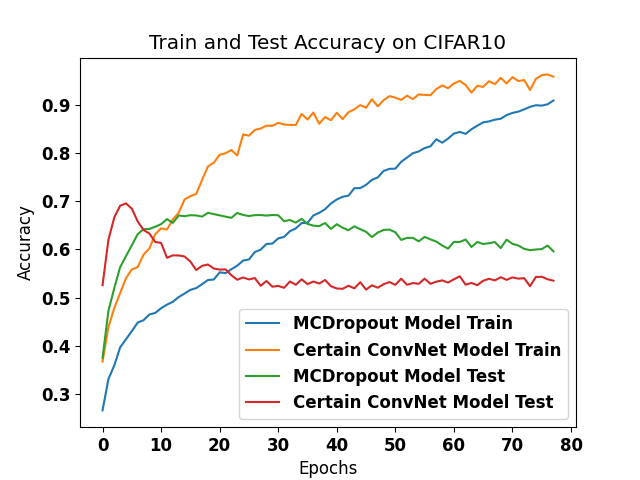} &
      \includegraphics[width=0.3\linewidth]{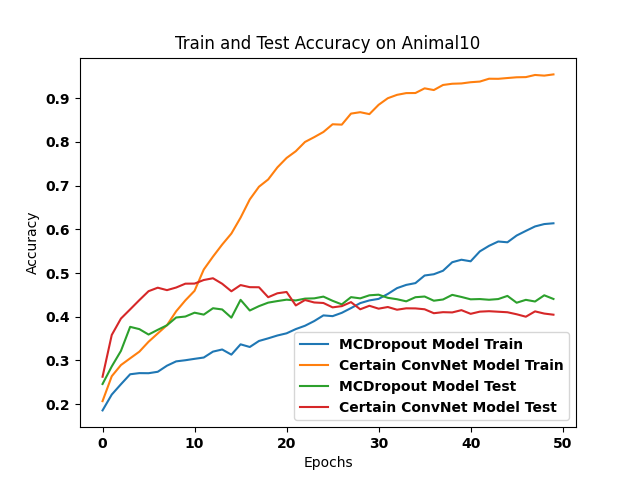} 
      \\
  \end{tabular}
    \caption{We show the train and test accuracies per epoch for (Left) the  certain and MCDropout LeNet5 models on MNIST; (middle) the certain and MCDropout ConvNet models on CIFAR10; (right) the certain and MCDropout ConvNet models on Animal-10n. All of these models are trained in a noisy-label setting with 35 percent noise. In all these cases, the MCDropout model has a better validation accuracy than the certain model; the certain model generally overfits to training set noise. }
    \label{fig:train_test_accs}
\end{figure*}

\subsection{Measuring Efficacy }

We first train MCDropout and certainty models on training data with noisy labels and evaluate their accuracy on a cleanly labeled test set. We present the learning behaviors during training and testing over epochs.

\subsection{Measuring Responsiveness }
Next we compare neuron responsiveness measured by volatility in the two models. We define volatility as the standard deviation of a neuron's activations over a dataset; if a neuron is capable of producing vastly different activation values for different input images, the neuron's activation gamut would possess high standard deviation, indicating a highly responsive neuron.

To compute the activation gamut of a neuron, we first cache the feature maps $u_{j}^i$ , post-ReLu, produced by the $i$-th neuron on the $j$-th test set image. We find the mean activation value, $a_{i}^j$ for the feature map $u_{j}^i$.  In other words, for a feature map with $n$ rows and $m$ columns,

\[ a_{i}^j = \frac{\sum_{r=0}^{n} \sum_{c=0}^{m} u_{i}^j(r, c)}{nm}. \]

Per neuron, this results in $j$ values which compose its activation gamut $A_{i}$ 

\[ A_{i} = \{a_{i}^0, a_{i}^1 ... a_{i}^j \}. \]

We can perform statistical analysis on these gamuts and aggregate them per-layer, such as finding the mean activation value $V_{l}$ of all $I$ neurons in the $l$-th network layer:

\[ V_{l} = \frac{\sum_{i=0}^I mean(A_{i})}{I}.\]

We also find the average gamut standard deviation  $S_{l}$ for all $I$ neurons in the $l$-th network layer 

\[ S_{l} = \frac{\sum_{i=0}^I std(A_{i})}{I}.\]

We would observe the activation gamut of a volatile neuron to possess a higher standard deviation than that of a nonvolatile neuron. The activation gamut of a volatile neuron may also include extremes, showing a higher maximum activation than a  non-volatile neuron. 

\subsection{Measuring Sparsity}
%\subsection{Network Sparsity on a Per-Neuron Level}

Along with research directions into network uncertainty and robustness, neural network sparsity has become a subject of interest for many machine learning researchers ~\cite{Gale2019TheSO,pruning,sparsetraining}. Sparse neural networks are desirable because they require less computation at test time, demand less memory~\cite{Gale2019TheSO}, and are less likely to overfit to training data~\cite{sparsitysurvey}. In the context of our investigation, the tendency for sparse neural networks to overfit more slowly to training data can allow them to avoid memorizing noisy training labels. We can evaluate network sparsity on a per-neuron level: which neurons never or rarely activate, for any and all test samples, and how common are these neurons throughout the entire model? Network sparsity can be defined as the subset of neurons output a value that is always zero ~\cite{Gale2019TheSO}, or very close: these neurons do not affect the final predictions in any significant manner. The larger the subset of neurons with this property, the more sparse the network's learned representation is. Because neural networks can easily overfit to noise in training labels~\cite{chen2021}, we are interested in the observed property of sparse models to overfit more slowly. With fewer tunable parameters available, sparse models have fewer degrees of freedom to overfit to noise.

\section{Results} \label{sec:experiments}
%Experimental set up

\begin{table*}%[ht!]
    \centering
    \begin{tabular}{|l|c|c|c|c|c|c|c|}
    \hline
        Metric & Model & conv0 & conv1 & fc1 & fc2 & fc3 \\ \hline
         Activation STD & Certain MNIST & 0.215 & 0.5367 & 2.386 & 1.335 & 1.5733   \\
         & MCDropout MNIST  & \textbf{0.0646} & \textbf{0.1085} &  \textbf{0.4207}  & \textbf{0.2144} & \textbf{0.9182} \\  \hline
         Activation Mean & Certain MNIST & 1.009 & 1.3936 & 1.4381 & 0.8383 & -0.0324  \\
         & MCDropout MNIST & \textbf{0.2443} &  \textbf{0.2041} & \textbf{0.1567} & \textbf{0.1208} & \textbf{-0.498} \\ \hline
         Unresponsive neurons & Certain MNIST & 0.0 & 0.0 & 0.0916 & 0.0 & 0.0  \\ 
         & MCDropout MNIST & \textbf{0.1666} & \textbf{0.25} & \textbf{0.5083} & \textbf{0.2023} & 0.0 \\ \hline
         
    \end{tabular}
    \caption{
    Quantitative metrics for our MNIST experiment, with the same table layout as described in Table \ref{tab:quantitative_cifar}. For all layers, the certain model's activations posses higher means and standard deviations, while the uncertain model has more relatively unresponsive neurons and lower activation values. Notice how several of the layers in the certain model have \emph{no} relatively unresponsive neurons.}
    \label{tab:quantitative_mnist}
\end{table*}

\begin{table*}%[ht!]
    \centering
    \begin{tabular}{|l|c|c|c|c|c|c|c|c|c|}
    \hline
        Metric & Model & conv0 & conv1 & conv2 & conv3 & fc1 & fc2 & fc3 \\ \hline
         Activation STD & Certain ConvNet & 0.0602 & 0.0343 & 0.1715 & 0.1279 & 7.3578 & 11.8364 & 6.5248  \\
         & MCDropout ConvNet  & \textbf{0.047} & \textbf{0.0123} & \textbf{0.0708} & \textbf{0.0871}  & \textbf{4.3155} & \textbf{9.2449} & \textbf{4.480} \\  \hline
         Activation Mean & Certain ConvNet & 0.0818 & 0.04378 & 0.238 & 0.106 & 1.6077 & 5.038 & 0.075  \\
         & MCDropout ConvNet & \textbf{0.060} &  \textbf{0.0149} & \textbf{0.091} & \textbf{0.0616} & \textbf{0.4894} & \textbf{3.340} & \textbf{-2.424} \\ \hline
         Unresponsive neurons & Certain ConvNet & 0.4166 & 0.4583 & 0.4323 & 0.4414 & 0.5449 & \textbf{0.2031} & 0.0 \\
         & MCDropout ConvNet & \textbf{0.4583} & \textbf{0.71875} & \textbf{0.7083} & \textbf{0.6172} & \textbf{0.7851} & 0.0781 & 0.0  \\ \hline
         
\hline
    \end{tabular}
    \caption{Quantitative metrics for our CIFAR10 experiment. We show the standard deviation of the mean activation value per neuron (top), the average of the mean activation value per neuron (middle), and a ratio of the ConvNet layer's neurons that are relatively unresponsive (bottom). We accumulate these metrics  across all test samples. In general, the certain model's layer activations have a higher standard deviation and higher mean (suggesting greater volatility). The uncertain model is less volatile and has more unresponsive neurons (suggesting sparsity).}
    \label{tab:quantitative_cifar}
\end{table*}

\begin{table*}%[ht!]
    \centering
    \begin{tabular}{|l|c|c|c|c|c|c|c|c|c|}
    \hline
        Metric & Model & conv0 & conv1 & conv2 & conv3 & fc1 & fc2 & fc3 \\ \hline
         Activation STD & Certain ConvNet & 0.2037 & 0.0202 & 0.0596 & 0.0304 & 2.077 & 5.071 & 5.695 \\
         & MCDropout ConvNet  & \textbf{0.0191} & \textbf{0.0132} & \textbf{0.0277} & \textbf{0.0301} & \textbf{1.6326} & \textbf{4.032} & \textbf{3.8372} \\  \hline
         Activation Mean & Certain ConvNet & 0.0217 & 0.0146 & 0.0599 & 0.0191 & 0.4833 & 3.017 & -2.256  \\
         & MCDropout ConvNet & \textbf{0.0172} & \textbf{0.0086} & \textbf{0.0265} & \textbf{0.0148} & \textbf{0.210} & \textbf{1.534} & \textbf{-1.9296} \\ \hline
         Unresponsive neurons & Certain ConvNet & 0.625 & \textbf{ 0.6354} & 0.4583 & \textbf{0.6367} & 0.4804 & 0.1094 & 0.0 \\
         & MCDropout ConvNet & \textbf{0.6666} & \textbf{0.7708} & \textbf{0.7448} & 0.4687 & \textbf{0.6679} & \textbf{0.5} & 0.0 \\ \hline
         
\hline
    \end{tabular}
    \caption{Quantitative metrics for our Animal-10n experiment, with the same table layout as described in Table \ref{tab:quantitative_cifar}. While less exaggerated than results shown on MNIST in Table \ref{tab:quantitative_mnist} and CIFAR in Table \ref{tab:quantitative_cifar}, we see a similar trend. For the majority of layers, the certain model is more volatile, with a larger gamut of possible activation values and higher overall magnitude of activations, while the uncertain model has more relatively unresponsive neurons and a sparser representation.
    }
    \label{tab:quantitative_animal10}
\end{table*}

We study the classification efficacy, neuron responsiveness, and network sparsity of MCDropout and certainty model on three benchmark classification datasets: MNIST, CIFAR10, and Animal-10n \cite{song2019selfie}. We use two different architectures: LeNet5 \cite{726791} and ConvNet, a convolutional neural network architecture with 4 convolutional layers followed by 3 fully connected layers. %Our investigation compares findings for a certainty model architecture and 
To maximize the effect of MCDropout, we use an all-layer MCDropout architecture where each layer in the certainty model is augmented with MCDropout. Our investigation compares the findings for the original certainty model and its augmented all-layer MCDropout model. 

We train both models on noisy training labels. Because we are evaluating these models on a classification task, mislabeled data simply means that training samples labeled with the incorrect class. For MNIST and CIFAR10, we use a uniform noise simulation scheme to add noise to 35\% of our training labels: in this scheme, each corrupted label has an equal chance of 35\% being mislabeled as any of the other classes. Once training is complete, we run our trained models on clean test data and compute all the neurons' individual activation maps after the application of an activation function. All of our chosen architectures use ReLu as their activation.

\subsection{Classification Efficacy}

We compare the performance of the certainty model and MCDropout models trained on noisy data and plot their training and accuracy curves over time \ref{fig:train_test_accs}. We see an emerging trend consistent across all models and datasets. The certain model overfits to the noise in the training data and results in a similar or higher final training accuracy than the MCDropout model. However, the MCDropout model consistently produces a higher validation accuracy.

Next we investigate why MCDropout outperforms certainty by analyzing the representations learned by both models. Consider the results on MNIST shown in \ref{fig:train_test_accs}, left. Given that the training accuracies of the certainty and MCDropout model are quite similar after 100 epochs, both models are clearly learning \emph{something}. However given the vastly different test accuracies between the two models--the uncertain models undoubtedly generalize better to the test dataset--the models are representing information \emph{differently}.

\subsection{Neuron Responsiveness Measured by Volatility}

Next, we compare the volatility of the neurons in the uncertain and certain models. We do so with two strategies. As mentioned earlier, we cache each neuron's activation map for every image in the test set. We find the mean activation value for each feature map. To measure volatility, we compare two statistics per layer: the standard deviation of the layer's mean activation values and the mean of the layer's mean activation values.

We show the results of this investigation for each dataset in Table \ref{tab:quantitative_mnist} for MNIST, Table \ref{tab:quantitative_cifar} for CIFAR10 and Table \ref{tab:quantitative_animal10} for Animal-10n. With the exception of a few layers, the neurons in the certainty models activate more strongly: the mean activation is higher for each layer and with greater variation: the standard deviation of the activations is higher for each layer.

\begin{figure*}[t!]
    \centering
    \setlength{\tabcolsep}{1pt}
    \begin{tabular}{cc c c}
      
      \includegraphics[width=0.25\linewidth]{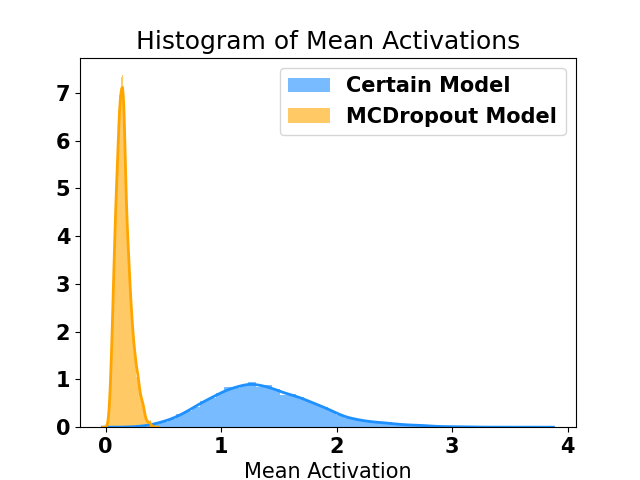} &
      \includegraphics[width=0.25\linewidth]{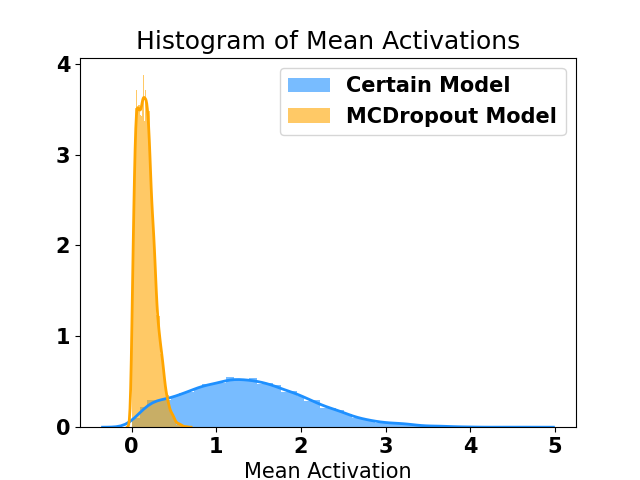} &
      \includegraphics[width=0.25\linewidth]{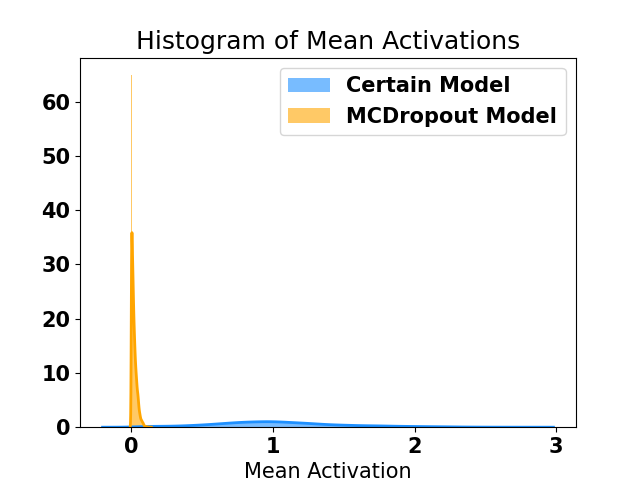} &
      \includegraphics[width=0.25\linewidth]{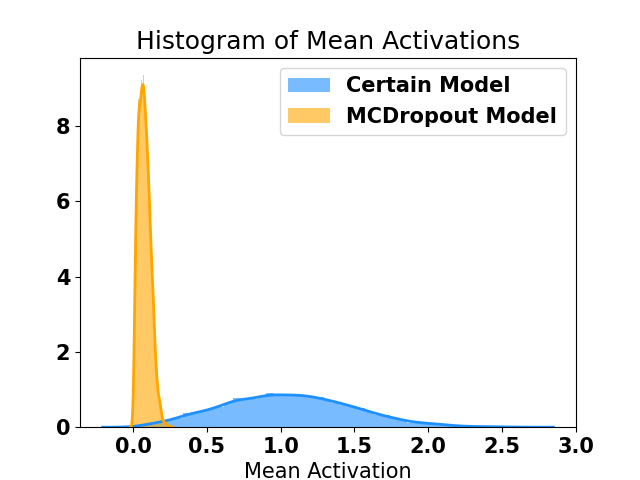} 
      \\
      \includegraphics[width=0.25\linewidth]{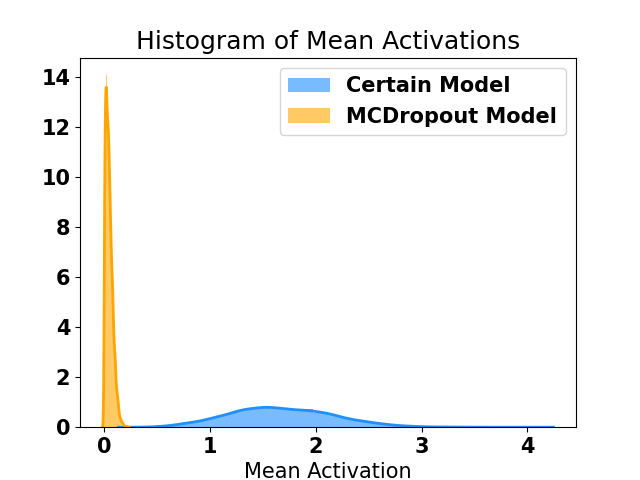} &
      \includegraphics[width=0.25\linewidth]{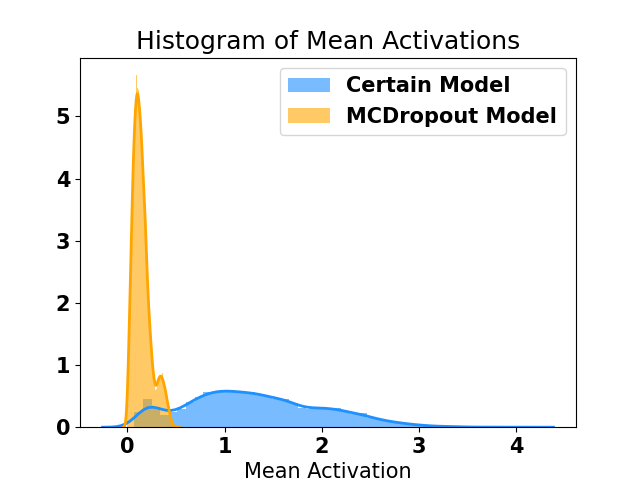} &
      \includegraphics[width=0.25\linewidth]{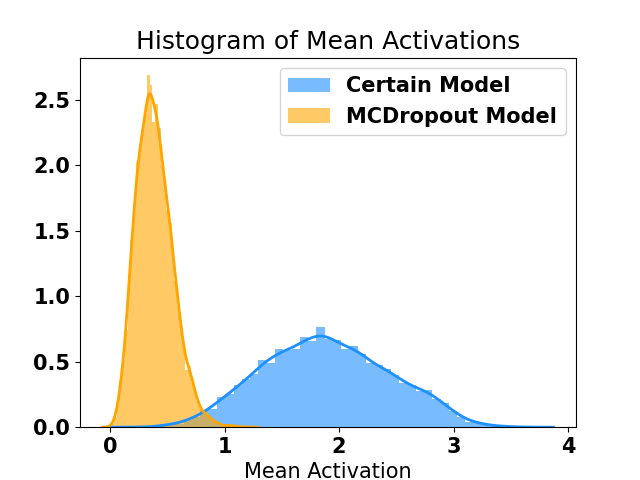} &
      \includegraphics[width=0.25\linewidth]{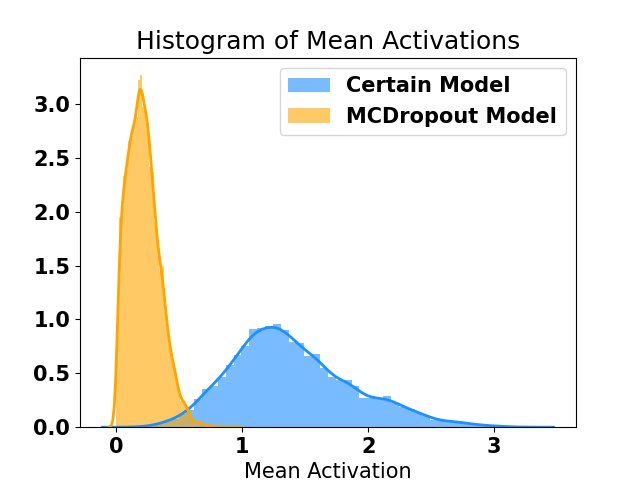} 
      \\
  \end{tabular}
    \caption{Histograms for our MNIST experiment, with the same set-up described in Figure \ref{fig:histogram_cifar}. We show 8 neurons from the second convolution layer. Several neurons from the uncertain model (orange) are unresponsive to all inputs: notice how many of the uncertain activation distributions are centered around 0, with shorter tails. This suggests that the uncertain model is sparser. We include a quantitative summary of activation values over the entire test dataset in Table \ref{tab:quantitative_mnist}
    }
    \label{fig:histogram_mnist}
\end{figure*}

\begin{figure*}[t!]
    \centering
    \setlength{\tabcolsep}{1pt}
    \begin{tabular}{cc c c}
      
      \includegraphics[width=0.25\linewidth]{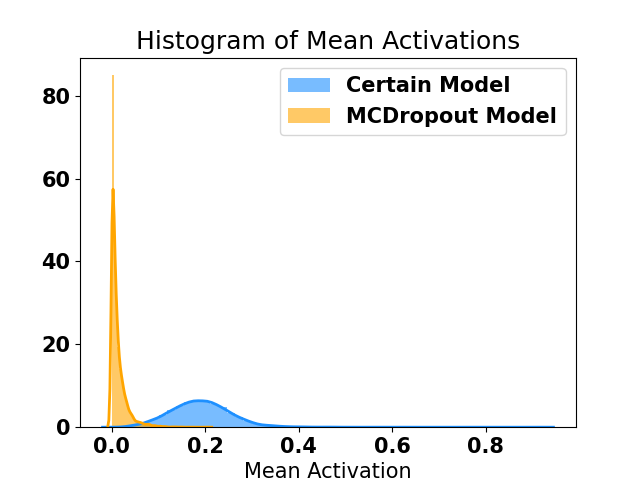} &
      \includegraphics[width=0.25\linewidth]{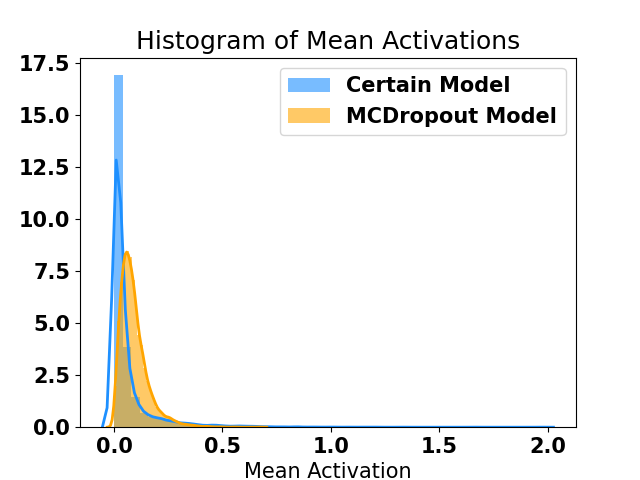} &
      \includegraphics[width=0.25\linewidth]{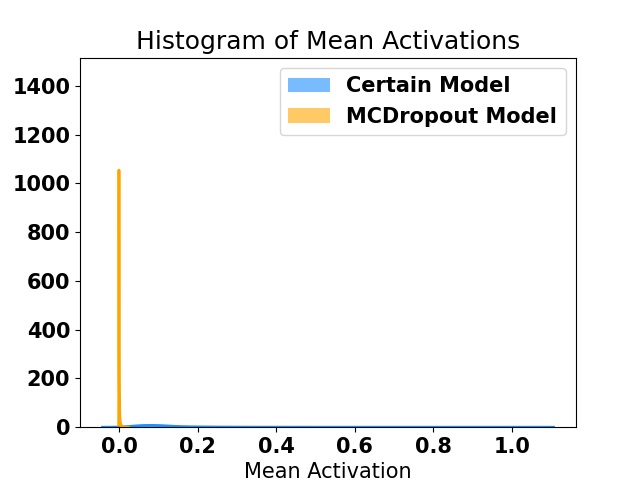} &
      \includegraphics[width=0.25\linewidth]{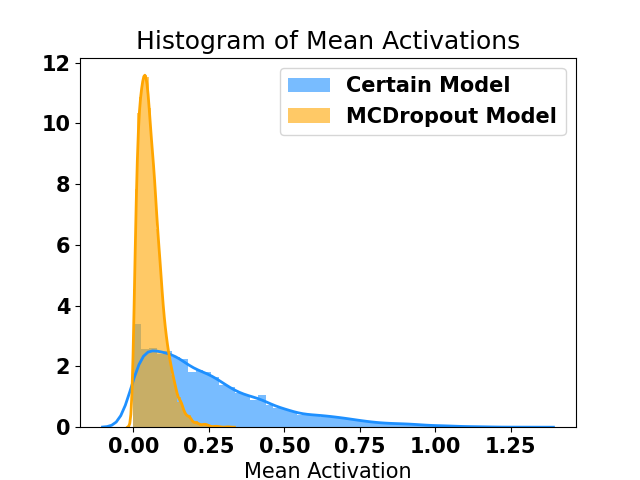} 
      \\
      \includegraphics[width=0.25\linewidth]{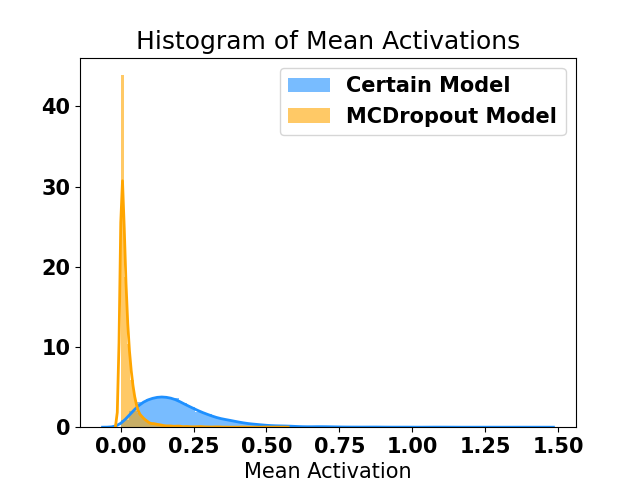} &
      \includegraphics[width=0.25\linewidth]{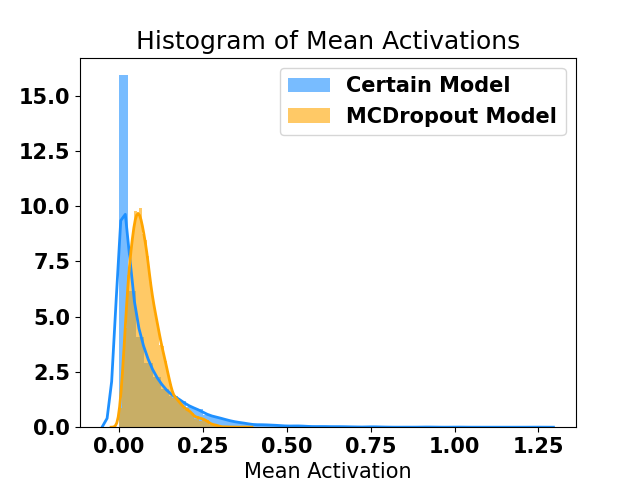} &
      \includegraphics[width=0.25\linewidth]{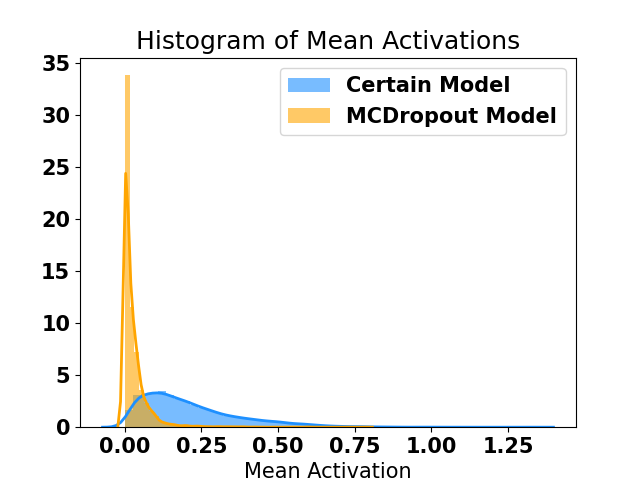} &
      \includegraphics[width=0.25\linewidth]{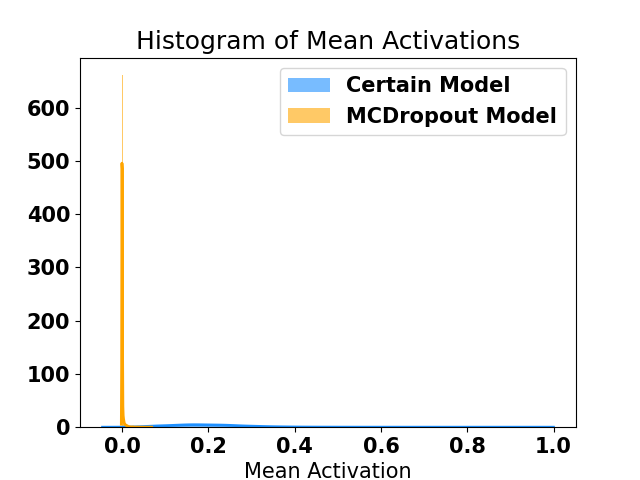} 
      \\
  \end{tabular}
    \caption{Histograms for our CIFAR10 experiment. We gather the mean activation values per neuron for each image in the CIFAR10 test set, for both certain and MCDropout models. We show 8 neurons from the 1st convolution layer; for each neuron, we plot a histogram of the mean activation values. Mean activations from the certain model are in blue; uncertain model are in orange. Notice how the range of the uncertain model's distributions are smaller: uncertain neurons produce a smaller gamut of possible mean activations. The uncertain model's distributions tend to be centered on or near 0. This suggests several of the uncertain model's neurons produce no activation no matter the input image. We choose 8 neurons to fit in page requirements; we include a quantitative summary of these metrics in Table  \ref{tab:quantitative_cifar}.
     }
    \label{fig:histogram_cifar}
\end{figure*}

\begin{figure*}[t!]
    \centering
    \setlength{\tabcolsep}{1pt}
    \begin{tabular}{cc c c}
      
      \includegraphics[width=0.25\linewidth]{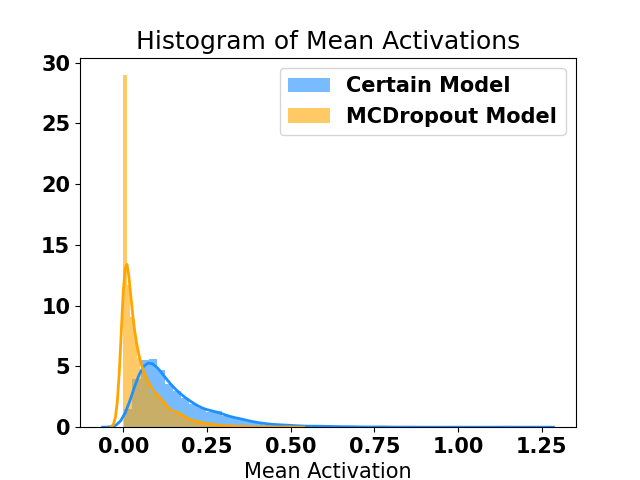} &
      \includegraphics[width=0.25\linewidth]{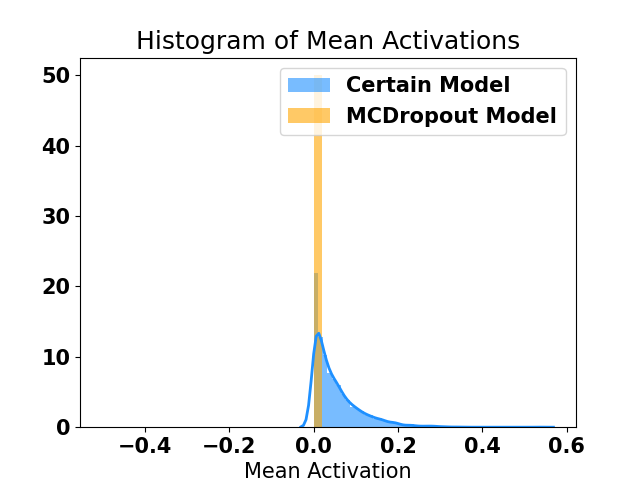}  &
      \includegraphics[width=0.25\linewidth]{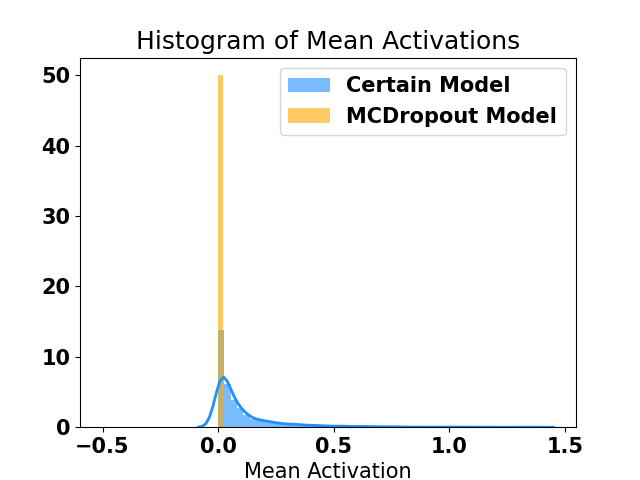}  &
      \includegraphics[width=0.25\linewidth]{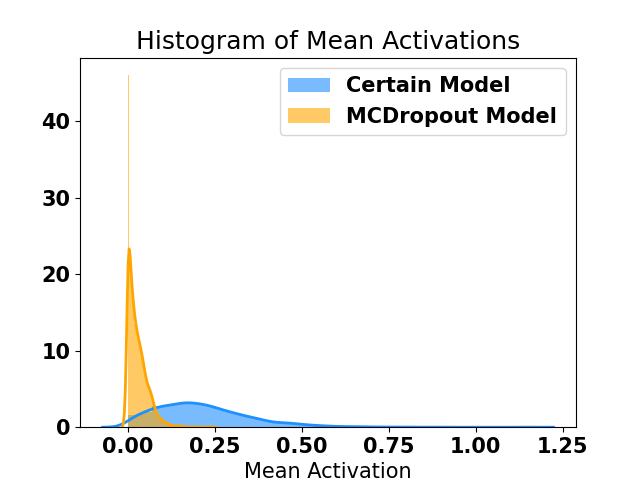}  
      \\
      \includegraphics[width=0.25\linewidth]{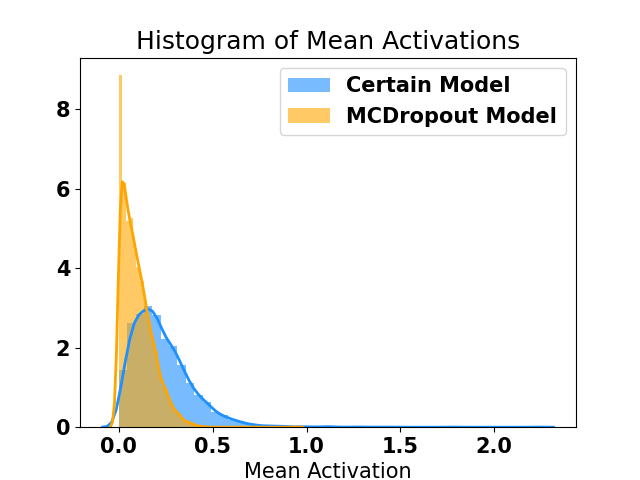}  &
      \includegraphics[width=0.25\linewidth]{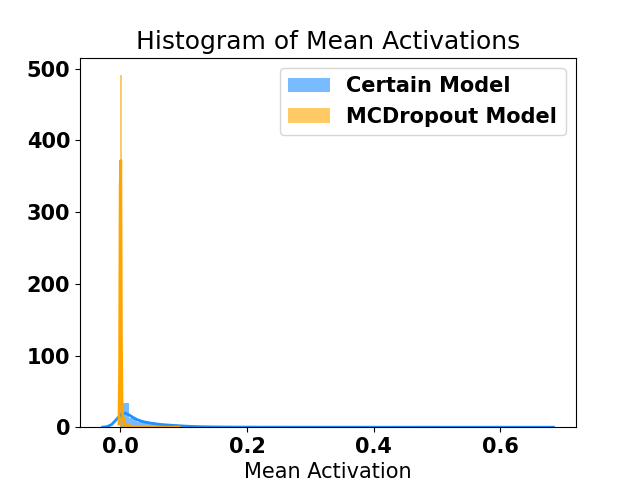}  &
      \includegraphics[width=0.25\linewidth]{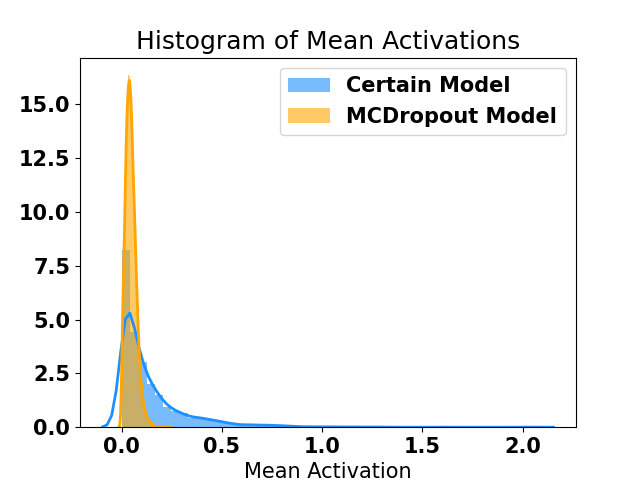} &
      \includegraphics[width=0.25\linewidth]{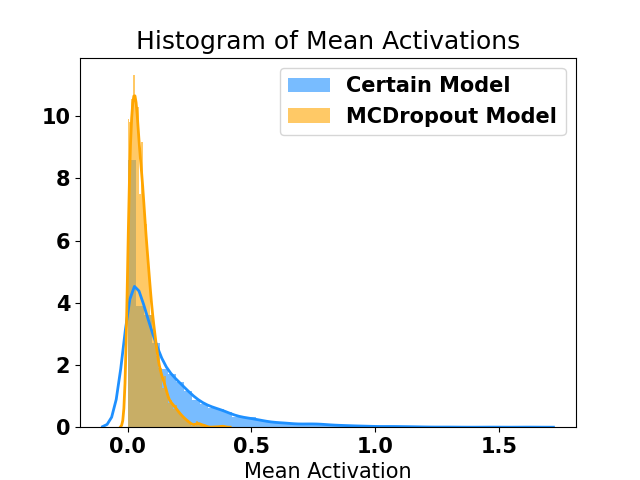} 
      \\
  \end{tabular}
    \caption{Histograms for our Animal-10n experiment, with the same set-up described in Figure \ref{fig:histogram_cifar}. We show neurons from the third convolution layer. Similar to Figure \ref{fig:histogram_cifar}, several neurons from the uncertain model are unresponsive (result in mean activations near 0) to all inputs. This suggests that the uncertain model is sparser.  We include a quantitative summary of activation values over the entire dataset in Table \ref{tab:quantitative_animal10}}.
    \label{fig:histogram_animal10}
\end{figure*}

\subsection{Network Sparsity}

We can compare the sparsity via analysis of the neurons' individual feature maps. Sparser models contain more neurons whose feature maps have values close to some constant \textbf{c}, usually 0, no matter the input sample from the test set. 

For qualitative evaluation, we visualize the post-activation feature map of individual neuron for a given sample of testing data. We can visually compare how many neurons seem near constant or activate in only small patches of the map. We show heatmaps from various layers in Figure \ref{fig:heatmap_mnist} (MNIST), Figure \ref{fig:heatmap_cifar} (CIFAR10), and Figure \ref{fig:heatmap_animal10} (Animal-10n). In all cases, activation maps from MCDropout models have spatially sparse activations: when they do activate, it is in tight, localized regions, and large patches of each activation map remain inactivated. In addition, several of the MCDropout activation maps show very little activation at all. 
We also calculate the mean activation value per neuron per image in each experiment's test dataset. We plot per-neuron histograms of these mean activation values for the certain and MCDropout models in Figures \ref{fig:histogram_mnist} (MNIST), \ref{fig:histogram_cifar} (CIFAR10), and \ref{fig:histogram_animal10} (Animal-10n). We can then compare the mean and support of the resulting activation distributions: in many cases, the distributions from MCDropout models possess smaller supports and are centered more closely around a mean activation value of 0.0. This provides an intuitive understanding of why MCDropout is more robust against noisy labels: the neurons that may be influenced by noisy labels in the certainty model are not activated in MCDropout models. MCDropout layers provide regularization against these ``corrupted" neurons.

%many neurons of the MCDropout model activate rarely over the entire test set. The same is not true for the certainty models.

These qualitative traits show that the MCDropout model's learned representation is more sparse. For a quantitative analysis, we can count how many neurons are ``relatively unresponsive" based on their gamut of possible activations for all the test images. Neurons that rarely activate--that is, the mean of their activations for all images on the test set falls below some epsilon threshold--are tallied in the final row of Tables \ref{tab:quantitative_mnist} (MNIST), \ref{tab:quantitative_cifar} (CIFAR10), and \ref{tab:quantitative_animal10} (Animal-10n). We report these numbers as the ratio of ``relatively unresponsive neurons" to the total number of neurons in the layer. The results show that the major of the MCDropout models' layers have more dead neurons than corresponding layers in the certain model does. This indicates that the uncertain model has learned a more sparse representation.

% \subsection{Informed Tuning of MC-Dropout Hyperparameters}

% In the next experiment, we consider: what property of MCDropout makes an uncertain network more sparse and less volatile, and how can this property inform how we choose network parameters? The hyperparameter we focus on is the \emph{number} of layers in a model that we convert from a normal layer into an MCDropout layer.

% We hypothesize that MC-Dropout layers have sparser, less volatile representations because of the regularization of dropout. Dropout has been shown to be a strong regularizer; the combination of several layers of dropout and the ensemble nature of MC-Dropout means that there is much more dropout to go around.

% If this is the case, the hypothesis suggests that better accuracies will be retrieved by simply using more MCDropout layers. 

\section{Discussion} \label{sec:discussion}

We have compared the representations, on a per-neuron level, learned by MCDropout models and certainty models when trained with noisy labels. The representation learned by MCDropout representation is less volatile but more sparse, an apt justification for its greater effectiveness and generalization in noisy-label scenarios. MCDropout provides regularization so that neurons are not overly influenced by the noisy labels; as a result, these neurons are not activated at test time, thus contributing to the robustness against noisy label training. With fewer free parameters to over-explain training label noise, MCDropout models forge representations that are less capable of overfitting to noisy labels.

Our larger goal in this investigation is not to build state-of-the-art models on any of the presented datasets or find the best network to deal with noisy labels. Rather, we investigate interpretable metrics and observations from the learned representations of models that identify why MCDropout model outperforms certainty model.

%%%%%

\begin{figure}
    \centering
    \includegraphics[width=0.8\linewidth]{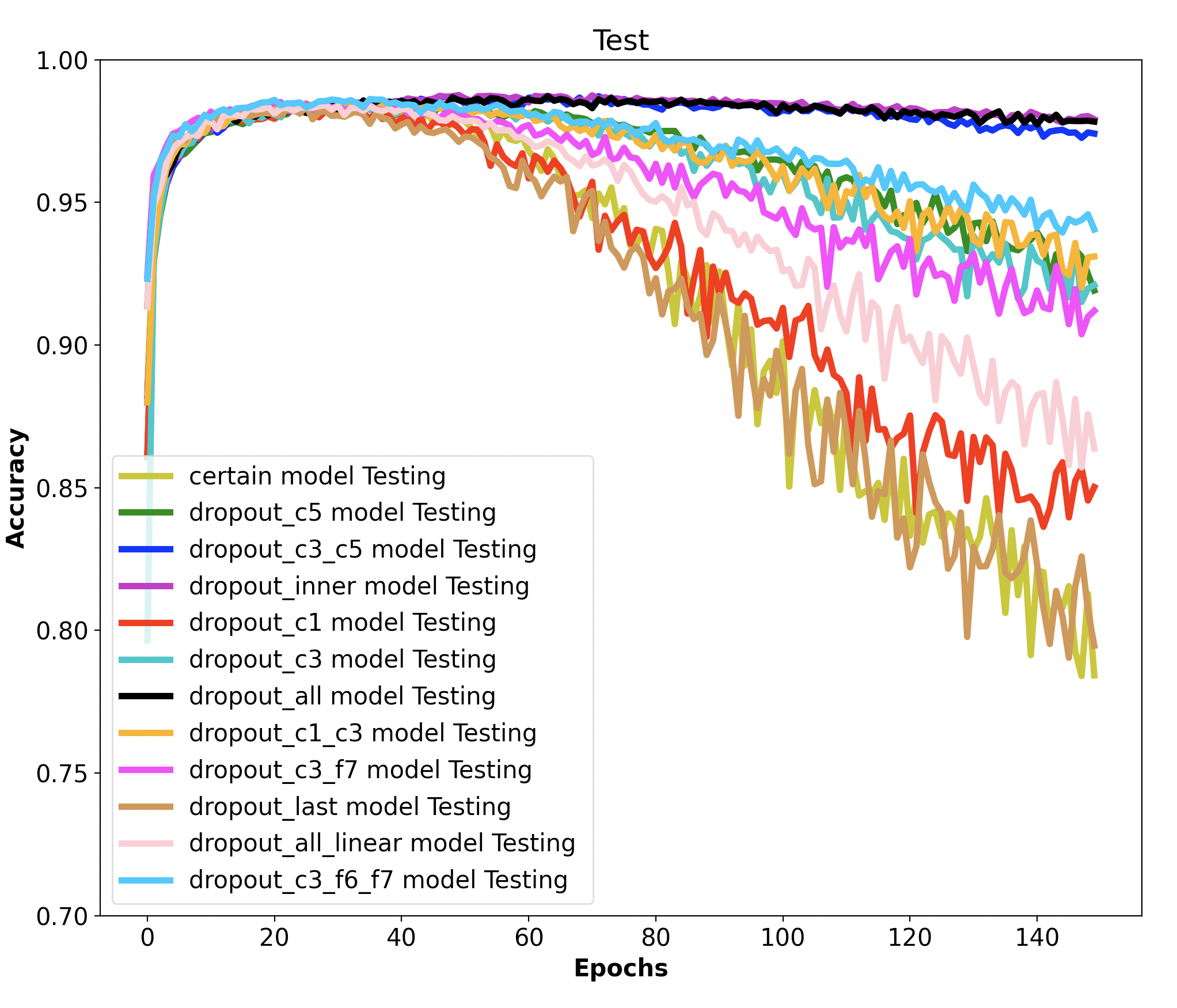}
    \caption{We further analyze different MCDropout placement configurations and discover that MCDropout on all layers (black line) possesses the optimal results against noisy labels, particularly compared to the certain model (yellow line).  }
    \label{fig:uncertainty_placement}
\end{figure}

In our experiment, we primarily analyzed all-layered MCDropout for the purpose of maximizing the MCDropout effect. However we acknowledge there are other different configurations of uncertainty placement. 
As seen in Figure \ref{fig:uncertainty_placement}, we further analyze different MCDropout placement configurations on MNIST dataset and discover that MCDropout on all layers possesses the best test classification accuracy when training with noisy labels. Such behavior is consistent with the theoretical establishment that all-layered MCDropout best approximates Bayesian neural network. While other configurations such as converting only convolutional layers, internal layers, final layers, etc., to MCDropout layers still outperform the certainty model, the best-performing model benefits from the most number of MCDropout layers. We believe research directions on an optimal trade off between classification performance and MCDropout layer placement is critical for noisy-label training with constraints on memory or inference time.

%%%%%

We hope this investigation helps us ask and answer more questions. It would be interesting to see if our observations about volatility and sparsity hold for other uncertainty estimation or ensemble methods like Bootstrap~\cite{bootstrap} or Bayes by Backprop~\cite{10.5555/3045118.3045290}. These observations may also help us tune MCDropout-related hyperparameters, such as the best locations to place MCDropout layer--or layers capable of uncertainty estimation in general--in a model architecture.

{\small
\bibliographystyle{ieee_fullname}
\bibliography{egbib}
}

\end{document}

%% file: latex/prelim.tex
\section{Preliminaries} \label{sec:prelim}
In this section, we present the problem statement, the preliminaries on label noise, Monte Carlo Dropout and related work. 

We consider a fully supervised learning problem in image classification, where the images and its associated labels in the training set, denoted by $ \mathcal{T}_{train} := \{(X_i, Y_i)\}_{i=1}^{n}$, with $n$ denoting the total number of training samples and all the  pairs $\{(X_i, Y_i)\}_{i=1}^n$ sampled i.i.d from a joint distribution $F_{X,Y}$.  However instead of observing all the correctly annotated labels, we observe the training data  $ \mathcal{T}_{tr} :=  \{(X_i, \hat{Y}_i)\}_{i=1}^n$, where given by a probabilistic process, $\hat{Y}_i$ deviates from $Y_i$. Our exploitation task is to learn a robust classifier on $ \mathcal{T}_{tr} =  \{(X_i, \hat{Y}_i)\}_{i=1}^n$ containing noisy labels such that the classification efficacy on incoming test image $X$ can best predict the unknown label $Y$.

Across this paper, we refer to the deterministic neural network without uncertainty estimation as \textit{certainty model} or \textit{deterministic model" interchangeably}. We  refer to the neural network augmented with  MCDropout layers as the \textit{MCDropout model}. 

\subsection{Label Noise Taxonomies}
There are several categorizations of noise labels. One commonly used categorization depends on whether or not the noisy label depends on the features. If the noisy label generation process is conditionally independent of the features, then a noise transition matrix $T_{c \times c}$, where $c$ is the number of classes, is sufficient to describe the label noise generation process. Each entry in $T_{ij} = p_{ij}$ is a probability such that the true label will be changed into a noisy label with probability $p_{ij}$.  If the observed label is different from the true label with a uniform probability, then the noise is considered to be label-independent and this noise is called considered symmetric or uniform noise. If the observed label is changed from the true label with probabilities depending on the original ground truth, then the noise is label-dependent and called asymmetric noise.  On the other hand, if the corruption process depends on the features and labels, the label noise is called instance-dependent. A more recent study proposes a new but practical assumption within instance-dependent label noise, defined as part-dependent label noise, where the noise depends partially on an instance \cite{xia2020part}. 

Another perspective on label noise is via uncertainty characterization \cite{chen2021}. The noisy label generation process is probabilistic and random. Naturally uncertainty characterization comes into play. From the notion of deep learning uncertainty, the noise in the labels can be considered a type of aleatoric uncertainty, a measurement of the intrinsic and irreducible uncertainty within the data. Within aleatoric uncertainty, homoscedastic uncertainty is constant across the input while heteroscedastic uncertainty is dependent on the input. Hence if the noise transition matrix is a uniform or symmetric one, then the label noise can be considered homoscedastic; if it is label-dependent, then the label noise can be considered heteroscedastic. In recent noise simulation schemes, label noise is applied on samples that are more likely to be mislabeled given by pre-learned model \cite{algan2020label}. We consider such type of noise as epistemic uncertainty, a term that describes uncertainty induced by models. 

% \subsection{Deep Learning Uncertainty}
\subsection{Monte Carlo Dropout}
The deep learning uncertainty perspective to characterize label noise inspires us to study label noise via deep learning uncertainty estimation techniques. Chen et al \cite{chen2021} proposed using epistemic uncertainty estimation methods when learning with noisy labels. Comparing Monte Carlo Dropout, Bootstrap \cite{bootstrap}, Bayesian CNN upon Bayes by Backprop \cite{10.5555/3045118.3045290} and certainty neural networks trained in noisy label settings, the authors discovered that Monte Carlo Dropout (MCDropout) had a prolonged memorization effect and possessed the best classification performance on test set.

\begin{figure}
    \centering
    \includegraphics[width=0.42\linewidth]{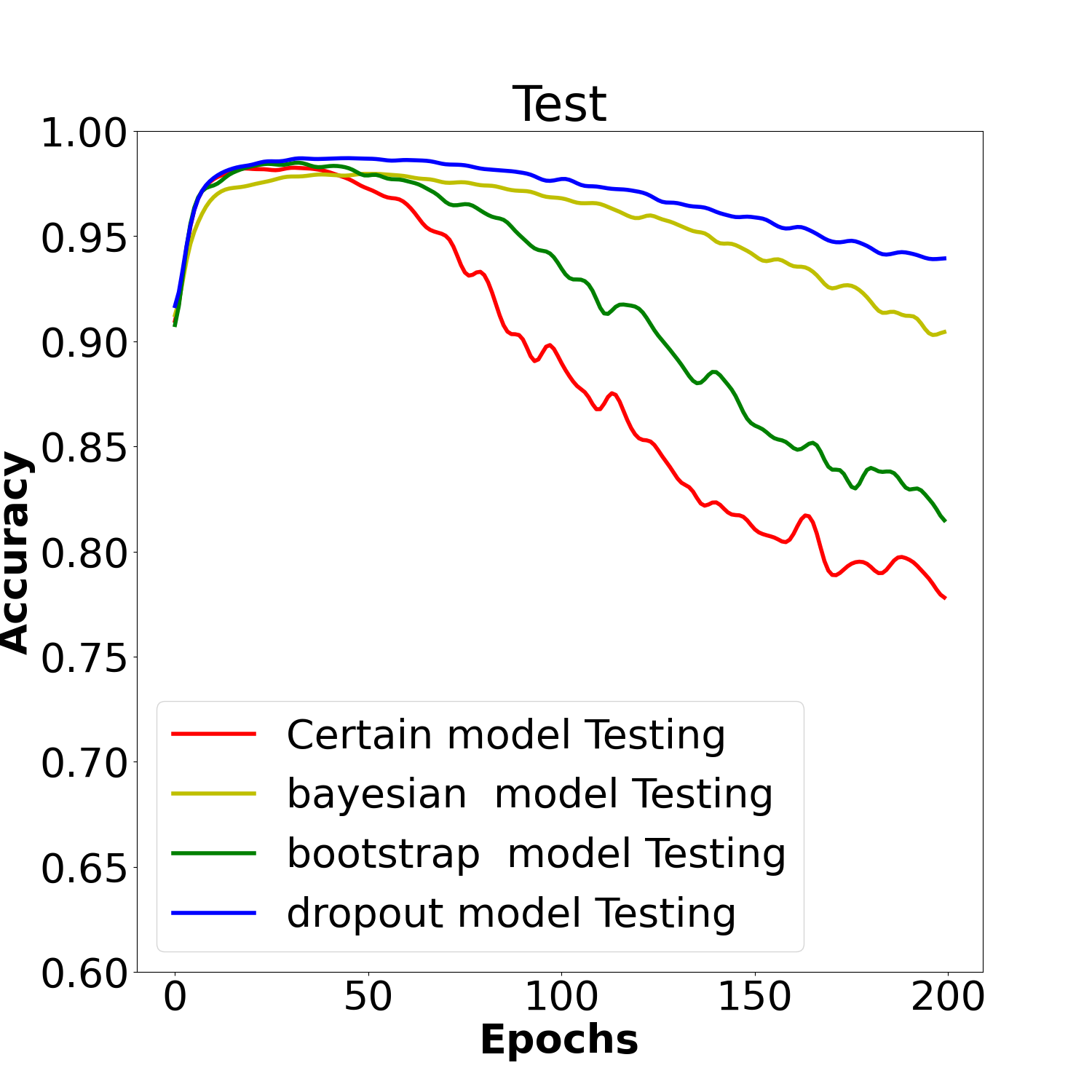}
    \includegraphics[width=0.42\linewidth]{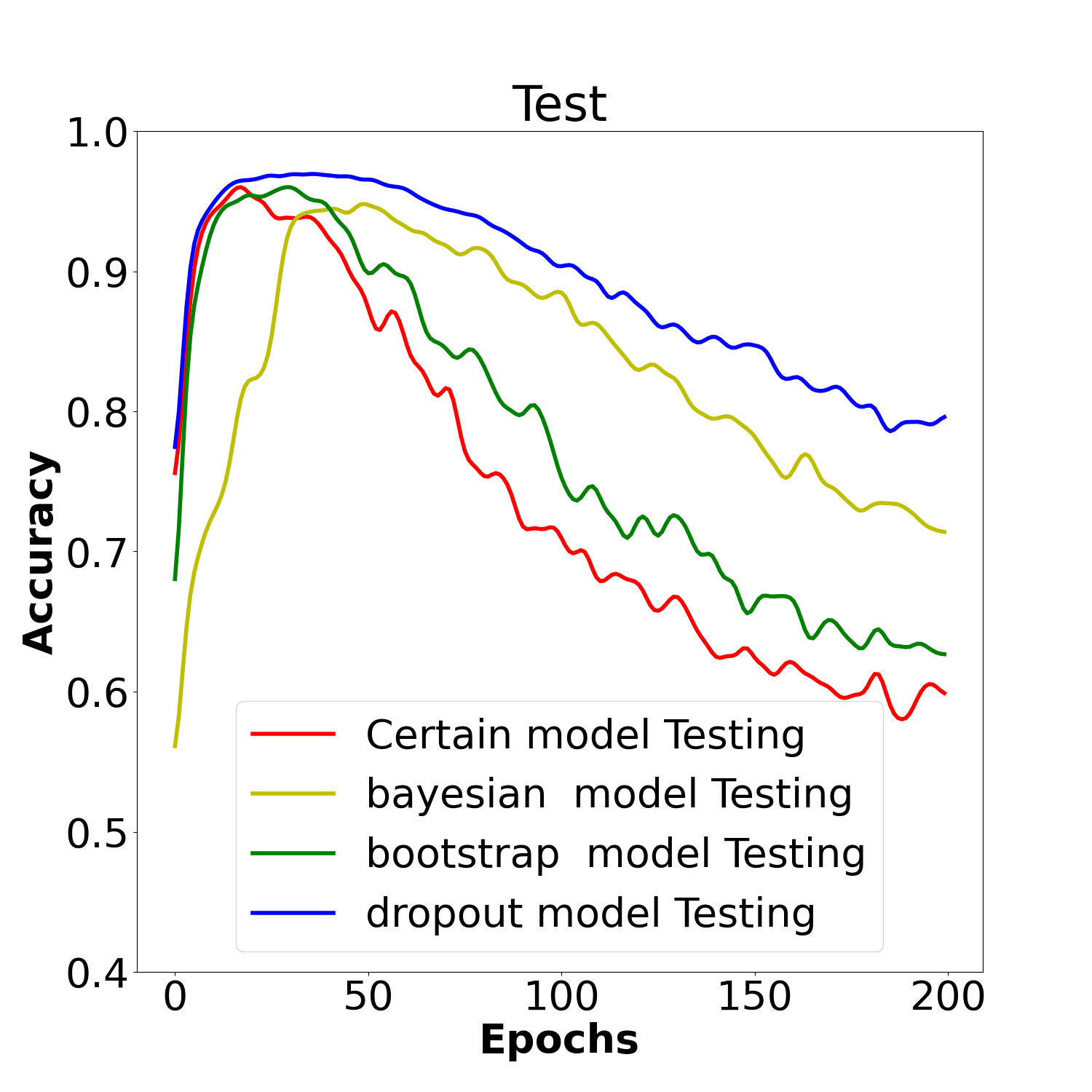}
    \caption{(Left): MNIST test accuracy when training labels contain 15\% noise. (Right): MNIST test accuracy when training labels contain 35\% noise. Our previous study suggests that MCDropout has the best classification performance among a few other uncertainty estimation methods. Further MCDropout does not increase training time per epoch and has relatively cheap inference cost. Hence in this paper, we focus on investigating the robustness of MCDropout when training with noisy labels.}
    \label{fig:motivation_examp}
\end{figure}

%%%%%%
We also included Figure \ref{fig:motivation_examp} as our motivational example here. Hence in this paper, we laser-focus on the study of why MCDropout possesses robustness against noisy labels in comparison with certainty models. In this section, we provide the background information on MCDropout.
%%%%%%

The core idea of MCDropout is to enable dropout regularization at both training and test time. With multiple forward passes at inference time, the prediction is not deterministic and can be used to estimate the posterior distribution. As a result, MCDropout offers Bayesian interpretation. First proposed in \cite{gal2016dropout}, the authors established the theoretical framework of 
MCDropout as approximate Bayesian inference and proved MCDropout minimises the Kullback–Leibler divergence
between an approximate distribution and the posterior of
a deep Gaussian process. More formally, let $d_l$ denotes dropout at the $l$-th layer of a neural network, where $d_l  \sim Bernoulli(p)$. Then at inference time, with $K$ forward passes, we obtained a distribution of $K$ logits and predictions per test data, where we can compute the expected value, standard deviation, variation ratio and entropy to assess uncertainty.

\subsection{Related Work on Deep Learning with Noisy Labels}
While there has not been much work on applying epistemic uncertainty methods to address noisy labels, an abundant of research has been done in deep learning noisy labels ranging from loss function adjustment, robust architecture design, data processing, data filtering and so on. Authors in \cite{ghosh2017robust, zhang2018generalized, wang2019symmetric, lyu2019curriculum} devised robust loss function to achieve a smaller risk for unseen clean data when learning with noisy labels. 
Sample selection techniques as to filter the clean labels for training and removing the noisy labels have been proposed in \cite{jiang2018mentornet, han2018co, yu2019does, malach2017decoupling}. Sample selection and label correction for spatial computing is studied in \cite{chen2020robust}.
Devising loss to estimate noise transition matrix and correct the labels are studied in \cite{patrini2017making, hendrycks2018using, arazo2019unsupervised}. Semi-supervised learning is another field of techniques on noisy labels, where the noisy labelled data are treated as unlabeled and clean labelled data are as labeled \cite{nguyen2019self, ding2018semi, li2020dividemix}. 

% Monte Carlo Dropout was first proposed in \cite{gal2016dropout}. 
% \subsection{A Motivational Example}
% In \cite{chen2021}, we presented a study on uncertainty estimation methods in the presence of noisy labels.  We compared classification performance of MCDropout, Bootstrap and Bayes by back-propagation against neural networks without uncertainty estimation and discovered that MCDropout has the best performance on test set when trained with noisy labels, as seen in Figure \ref{fig:motivation_examp}. Therefore, we focus our investigation on the properties of MCDropout models trained in noisy-label settings.